\documentclass{article}

    \usepackage[preprint]{neurips_2024}

\usepackage[utf8]{inputenc} 
\usepackage[T1]{fontenc}    
\usepackage{hyperref}       
\usepackage{url}            
\usepackage{booktabs}       
\usepackage{amsfonts}       
\usepackage{nicefrac}       
\usepackage{microtype}      
\usepackage{xcolor}         
\usepackage{amsmath}        
\usepackage{algorithm}      
\usepackage{algorithmic}    
\usepackage{graphicx}       
\usepackage{authblk}

\title{Subset Selection for Fine-Tuning: A Utility-Diversity Balanced Approach for Mathematical Domain Adaptation}

\author{
  Madhav Kotecha${^1}$, Vijendra Kumar Vaishya${^1}$, Smita Gautam${^1}$, Suraj Racha${^2}$ \\
  ${^1}$Department of Computer Science and Engineering, IIT Bombay\\
  ${^2}$Koita Centre for Digital Health, IIT Bombay\\
  \texttt{\{24m0833, 24m2133, 24m0845, 23d1627\}@iitb.ac.in}
}

\begin{document}

\maketitle

\begin{abstract}
We propose a refined approach to efficiently fine-tune large language models (LLMs) on specific domains like the mathematical domain by employing a budgeted subset selection method. Our approach combines utility and diversity metrics to select the most informative and representative training examples. The final goal is to achieve near-full dataset performance with meticulously selected datapoints from the entire dataset while significantly reducing computational cost and training time and achieving competitive performance as the full dataset. The utility metric incorporates both perplexity and Chain-of-Thought (CoT) loss to identify challenging examples that contribute most to model learning, while the diversity metric ensures broad coverage across mathematical subdomains. We evaluate our method on LLaMA-3 8B and Phi-3 models, comparing against several baseline approaches, including random selection, diversity-based sampling, and existing state-of-the-art subset selection techniques.
\end{abstract}

\section{Introduction}

Fine-tuning large language models (LLMs) on domain-specific tasks often requires substantial computational resources and large datasets. This is particularly intriguing in the mathematical domain, where problems vary widely in difficulty and structure. Current approaches either use the entire dataset (which is highly expensive in terms of computational expense) or select random subsets (potentially missing the important examples that have the highest gain). \\ 

We need the novel diversity-based subset selection that ensures that the subset selected is inclusive and representative of the entire large dataset. Diversity-based sampling methods such as the Determinant Point Process (DPP) help reduce redundancy but may not prioritize difficult or informative examples. \\

The proposed approach addresses these limitations by developing a selection strategy that balances both the informativeness and diversity of training examples, enabling more efficient fine-tuning while maintaining performance. The requirement of judicially selecting each data point arises as we progress toward the goal of subset selection, as every inclusion of a data point further adds up to the final gain we get by the particular generated subset. By applying this informed subset selection, we aim to achieve near-full dataset performance while reducing fine-tuning cost and carbon footprint. \\

The mathematics domain presents a particularly interesting test case due to its hierarchical nature, where understanding of fundamental concepts is often prerequisite to solving more complex problems and correctly learning the reasoning of one mathematical equation allows similar to be computed accordingly, the main approach that allows the data pruning of the entire dataset. 

\section{Background and Motivation}

The rapid advancement of LLMs demands the optimized solution for the existing problem that eventually decrease resources like time and cost. However, fine-tuning these models on domain-specific tasks presents significant challenges. The DELIFT \cite{agarwal2024delift} proposed solution designed to address data-selection across all stages of fine-tuning by the help of a single framework. It uses the intuition of in-context examples to get the information gain of each data point relative to the current state of the model. DELIFT \cite{agarwal2024delift} combines the pairwise utility with submodularity as proposed by SMART \cite{renduchintala2024smart}. The approach ensures that it selects a diverse, non-redundant subset uniquely curated for each phase of fine-tuning.

The main challenge that this study proposes to solve is as below:
\begin{itemize}
    \item \textbf{Computational Cost:} Fine-tuning small models and infer for the large model based on that is proposed by Small2Large \cite{yang2024smalltolarge}. It helps to reduce computation significantly by eliminating the requirement of performing the experiment on a large model to get the evaluation done for the selected method.
    \item \textbf{Dataset Size:} Mathematical datasets can be large and diverse, covering various topics and difficulty levels. The selection of data-points by consideration of the budget is proposed by \cite{banerjee2020budgeted} that also works under significant resource constraints
    \item \textbf{Example Importance:} Not all training examples contribute equally to model performance, with some being more informative than others. Prior work has shown that the some of the easiest examples from the dataset have no real contributions; eliminating some of the easiest examples can be more influential in teaching mathematical reasoning patterns.
    \item \textbf{Domain Complexity:} Mathematics encompasses numerous subdomains with different reasoning patterns and solution strategies, making it challenging to ensure comprehensive coverage with a small subset. The variant of the submodular function proposed by SMART \cite{renduchintala2024smart} allows us to eliminate the entire cluster of the example at once, which can be easily handle by some of the data-points of another nearby cluster.
    \item \textbf{Transfer Learning Efficiency:} Efficient transfer learning requires identifying examples that address gaps in the model's current knowledge rather than reinforcing already learned patterns.
\end{itemize}

Recent research in efficient fine-tuning has explored various approaches to address these challenges. Random subset selection, while computationally efficient, often results in suboptimal performance due to missing critical examples. Diversity-based methods like DPP ensure broad coverage but may include easy examples that contribute little to model improvement. Utility-based approaches select challenging examples but may overemphasize similar problem types, leading to poor generalization.

Our work is motivated by the observation that an ideal subset should balance both utility (selecting informative examples) and diversity (ensuring broad domain coverage). This is particularly important in mathematics, where understanding diverse problem-solving strategies is crucial for developing robust reasoning capabilities.

\section{Related Work}

\subsection{Efficient Fine-tuning Approaches}

Parameter-efficient fine-tuning methods like LoRA (Low-Rank Adaptation) \cite{hu2020lora} that focus on a small subset of trainable parameters instead of the entire model; similarly, prefix tuning has gained popularity for reducing memory requirements. However, these approaches still process the entire dataset, limiting computational savings. The proposed study is complementary to these methods, as our subset selection can be combined with parameter-efficient techniques for maximum efficiency.

\subsection{Data Selection for Fine-tuning}

Several approaches have been proposed for selecting subsets for fine-tuning:

\begin{itemize}
\item \textbf{Uncertainty-based Selection:} Methods like active learning select examples where the model has high uncertainty, but these can be computationally expensive to identify in large datasets. As the dataset size increases, the computation required to select highly uncertain examples also increases.
\item \textbf{Diversity-based Selection:} DPP and k-means clustering select diverse examples but may not prioritize informativeness, as diversity does not always focus on improving the model on highly challenging tasks.
\item \textbf{Hybrid Approaches:} Recent work by \cite{yang2024smalltolarge} proposed SmallToLarge (S2L), which uses a smaller model's training trajectory to identify valuable examples for larger models. The fine-tuning on smaller model is helpful to get the expected results and report for larger models.
\item \textbf{Submodular Optimization:} \cite{renduchintala2024smart} introduced SMART, a submodular data mixture strategy for instruction tuning that balances diversity and utility by considering that the gain by adding the element in the subset is always greater than or equal to the gain we get by adding the element in its superset.
\end{itemize}

Our approach builds upon these methods but is specifically tailored to the specialized domains like mathematics, incorporating domain-specific utility metrics like Chain-of-Thought loss that capture reasoning complexity.




\section{Proposed Approach}

We studied the subset selection method that uses utility value as the key metric to select the most valuable training examples out of all the examples present in the dataset. The implementation of a unified and computationally efficient data selection framework that works with multiple stages of fine-tuning and maximizes model performance while at the same time minimizing data redundancy. Further, we are using the idea of the approach for domain-specific data subset selection. While we proceed to the specific domain of the dataset, we need to analyze the additional features on top of the below-mentioned generalized approach. While diving into a deeper study of domains and characteristics associated with them, here we explained the mathematical domain to further specialize the approach for specific domains.

\subsection{Pairwise Utility Score}

Let $\mathcal{D} = \{(x_i, y_i)\}$ be a training set, where each $x_i$ is an input sequence/context and $y_i$ is the corresponding output. Consider two samples $(x_i, y_i)$ and $(x_j, y_j)$ that we want to compute the utility score, and let's define $G_i$ as the ideal ''ground truth'' distribution that assigns probability 1 to the token sequence $y_i$ and 0 otherwise. We define it $p(y_i \mid x_i)$ as the model’s predicted probability distribution of $y_i$ the given that $x_i$ alone and $p(y_i \mid x_i, x_j, y_j)$ as the predicted distribution of $y_i$ when it $(x_j, y_j)$ is also provided beforehand (e.g., as an in-context sample).

\subsection*{Utility Function for pair of data-points}
The DELIFT \cite{agarwal2024delift} calculates the \textit{information gain} $(x_j, y_j)$ for predicting $(x_i, y_i)$ via:
\begin{equation}\label{eq:4.1}
UF_{ij} = d(G_i, p(y_i \mid x_i)) - d(G_i, p(y_i \mid x_i, x_j, y_j))
\end{equation}
where $d(\cdot, \cdot)$ is a distance between probability distributions.

The length-normalized Euclidean distance that is used for numerical stability is as follows:
\begin{equation}\label{eq:4.2}
    d(GT_i,\, p(y_i \mid \cdot)) = \left\| 1 - p(y_i \mid \cdot) \right\|_2
\end{equation}

\subsection*{Informal Statement} \textit{If $d(\cdot, \cdot)$ is chosen to be the Kullback-Leibler (KL) divergence, then the utility $UF_{ij}$ coincides with the (conditional) pointwise mutual information between $y_i$ and $(x_j, y_j)$ given $x_i$. Formally,}
\begin{equation}\label{eq:4.3}
UF_{ij} = \log \frac{p(y_i \mid x_i, x_j, y_j)}{p(y_i \mid x_i)} = \sum_{t=1}^T \log \left( \frac{p(y_{it} \mid x_i, x_j, y_j, y_{i,<t})}{p(y_{it} \mid x_i, y_{i,<t})} \right)
\end{equation}

Thus, the utility function with $i$ and $j$ decides the information $(x_j, y_j)$ \textit{informs} the prediction of $y_i$ when given a $x_i$.

\subsection{Submodularity}
Let the function $f : 2^{\mathcal{V}} \rightarrow \mathbb{R}$ be a \textit{set function} that assigns a value to every subset of the \textit{ground set} $\mathcal{V}$. We consider the notation $f(v|X)$ = $f(X \cup \{v\}) - f(X)$, i.e., the incremental value \textit{gain} by adding the element $v \notin X$.

\subsection*{Submodular Function} A given set function $f : 2^{\mathcal{V}} \rightarrow \mathbb{R}$ is \textit{submodular} if for all $X, Y \subseteq \mathcal{V}$, where $X \subseteq Y$ and for all $v \notin Y$ and hence $v \notin X$, the following inequality holds true:
\begin{equation}\label{eq:4.4}
f(v|X) \geq f(v|Y) \quad \text{(diminishing gains property)}
\end{equation}

\textbf{Facility Location}\ref{appendix:submodfunctions} performs best when the requirment is to get diverse coverage of the instruction set among the available instructions. \textbf{Mutual Information}\ref{appendix:submodfunctions} works best to align samples with a particular target benchmark or domain $D_T$. \textbf{Conditional Gain}\ref{appendix:submodfunctions} to incorporate new, complementary data while avoiding redundancy with existing knowledge $D_E$ \\

\textbf{Subset Selection using Submodularity} \quad
To choose a subset $\mathcal{A}$ of size $k$, we apply a classic greedy heuristic~\cite{nemhauser1978analysis}, at each iteration we pick $d^* = \arg\max_{d \in \mathcal{D} \setminus \mathcal{A}} [f(\mathcal{A} \cup \{d\}) - f(\mathcal{A})]$. This yields a $1 - \frac{1}{e}$ approximation factor for submodular functions which is nearly \textbf{63.21\%}, ensuring that we obtain near-optimal subsets efficiently.

\begin{algorithm}[ht]
\caption{Greedy Maximization for Submodular Function}
\label{alg:1}
\begin{algorithmic}[1]
\REQUIRE Dataset $\mathcal{D}$, submodular function $f$, budget $k$
\STATE Initialize subset $\mathcal{A} \gets \emptyset$
\FOR{$t = 1$ to $k$}
    \STATE Select $d^* = \arg\max_{d \in \mathcal{D} \setminus \mathcal{A}} \left( f(\mathcal{A} \cup \{d\}) - f(\mathcal{A}) \right)$
    \STATE Update $\mathcal{A} \gets \mathcal{A} \cup \{d^*\}$
\ENDFOR
\STATE \RETURN $\mathcal{A}$
\end{algorithmic}
\end{algorithm}

\section{Mathematical Subset Selection}

We compute the utility score as a combination of Perplexity and Chain-of-Thought (CoT) Loss:

\begin{itemize}
    \item \textbf{Perplexity:} Measures how surprised the model is by the example, with higher values indicating more challenging examples. Formally, for an example with tokens $x_1, x_2, \ldots, x_n$, the perplexity is calculated as:
    \begin{equation}\label{eq:5.1}
        \text{PPL}(x) = \exp\left(-\frac{1}{n} \sum_{i=1}^{n} \log P(x_i \mid x_1, \ldots, x_{i-1}) \right)
    \end{equation}
    
    where $P(x_i \mid x_1, \ldots, x_{i-1})$ is the model's predicted probability of token $x_i$ given the preceding tokens.
   
    \item \textbf{CoT Loss:} Captures the model's difficulty in generating step-by-step reasoning, identifying problems that require complex reasoning. We compute this by comparing the model's generated reasoning steps with the ground truth reasoning:
    \begin{equation}\label{eq:5.2}
    \text{CoT Loss}(x) = \mathcal{L}(\text{CoT}_{\text{model}}(x), \text{CoT}_{\text{ground\_truth}}(x))
    \end{equation}
\end{itemize}

The combined utility score for an example $x$ is defined as:
\begin{equation}\label{eq:5.3}
    U(x) = \alpha \cdot \text{norm}(\text{PPL}(x)) + (1 - \alpha) \cdot \text{norm}(\text{CoT Loss}(x))
\end{equation}

where $\text{norm}(\cdot)$ normalizes the values to $[0,1]$ range, and $\alpha$ is a hyperparameter controlling the relative importance of perplexity versus CoT loss.

The utility score helps identify hard and informative samples that contribute significantly to model improvement.

\subsection{Diversity Score}

We calculate diversity using \textbf{cosine similarity} on sentence embeddings:

\begin{itemize}
    \item Generate embeddings for each mathematical problem using pre-trained sentence transformers such as SBERT or domain-specific mathematical embeddings.
    \item Compute pairwise cosine similarity between examples to create a similarity matrix $\mathbf{S}$ where 
    $S_{ij} = \text{cos\_sim}(e_i, e_j)$ for embeddings $e_i$ and $e_j$.
    
    \item Define the diversity score of a subset $\mathcal{A}$ as:
    \begin{equation}\label{eq:5.4}
        D(\mathcal{A}) = \sum_{i \in \mathcal{A}} \sum_{j \in \mathcal{A}, j \neq i} (1 - S_{ij})
    \end{equation}
    
    which rewards selecting examples that are dissimilar from each other.

\end{itemize}

\subsection{Subset Selection Algorithm}

Given a budget $\mathcal{B}$ (the maximum number of examples to select), our goal is to find a subset $\mathcal{A}$ of the full dataset that maximizes a combined objective function:
\begin{equation}\label{eq:5.5}
\max_{\mathcal{A} : |\mathcal{A}| \leq \mathcal{B}} \lambda \sum_{x \in \mathcal{A}} U(x) + (1 - \lambda) D(\mathcal{A})
\end{equation}
where $\lambda$ is a hyperparameter controlling the trade-off between utility and diversity.

This optimization problem is NP-hard, so we propose a greedy algorithm with theoretical guarantees:

\begin{algorithm}
\caption{Utility-Diversity Balanced Subset Selection}
\label{alg:2}
\begin{algorithmic}
\STATE \textbf{Input:} Dataset $X$, budget $B$, trade-off parameter $\lambda$
\STATE \textbf{Output:} Selected subset $A$
\STATE Compute utility scores $U(x)$ for all $x \in X$
\STATE Compute similarity matrix $S$ for all pairs in $X$
\STATE Initialize $A = \emptyset$
\WHILE{$|A| < B$}
\STATE $x^* = \arg\max_{x \in X \setminus A} \lambda U(x) + (1-\lambda) \sum_{y \in A} (1 - S_{xy})$ \ref{eq:5.5}
\STATE $A = A \cup {x^*}$
\ENDWHILE
\RETURN $A$
\end{algorithmic}
\end{algorithm}

The algorithm \ref{alg:2} has a submodularity property that guarantees the selected subset achieves at least $(1-1/e)$ the optimal solution's value. The calculated sub-optimal solution (that approximates to around > 63\% of the optimal solution) is further considered as representative of the particular domain/task to generalize over the cluster.

\subsection{Methodology Pipeline}

Our approach follows a systematic pipeline:

\begin{enumerate}

    \item \textbf{Dataset Preparation:} Collect and preprocess mathematical problems from various sources, ensuring coverage across different mathematical subdomains including algebra, calculus, geometry, number theory, and probability.
    
    \item \textbf{Utility Score Computation:} We first fine-tune a smaller proxy model (e.g., LLaMA-3 3B) on a small random subset and compute perplexity scores and generate CoT reasoning for each problem using this model. After computing CoT loss, we combine these metrics to form the utility score.
    
    \item \textbf{Diversity Score Computation:} We define the diversity function as generating embeddings for all mathematical problems and computing the similarity matrix.
    
    \item \textbf{Subset Selection:} After calculating both the score above, we apply the greedy algorithm \ref{alg:2} to select a subset that maximizes the combined utility-diversity objective.
    
    \item \textbf{Fine-tuning and Evaluation:} The final step is to fine-tune LLaMA-3 8B and Phi-3 models on the selected subset using appropriate hyperparameters and compare performance against baseline approaches on held-out test sets.
    
\end{enumerate}

\section{Experimental Setup}

\subsection{Datasets}

We are mainly using GSM8K and similar datasets like that, which are  publicly available mathematical problem datasets like GSM8K \cite{cobbe2021gsm8k}, MATH \cite{hendrycks2021math}, MMLU-Mathematics, and MathQA \cite{amini2019mathqa}. GSM8K is a dataset of 8,500 grade school math word problems requiring multi-step reasoning. MATH is a challenging dataset of 12,500 competition mathematics problems. MMLU-Mathematics is the mathematics subset of the Massive Multitask Language Understanding benchmark. MathQA is a large-scale dataset with 37,000 math word problems with step-by-step solutions.

For embedding generation, we will use MathBERT, a BERT model \cite{devlin2018bert} fine-tuned specifically on mathematical text, to better capture the semantic relationships between different mathematical problems.

\subsection{Models}

The models to evaluate our approach are two state-of-the-art LLMs, which are \textbf{LLaMA-3 8B \cite{touvron2023llama}} and \textbf{Phi-3 \cite{abdin2023phi}}. LLaMA-3 8B is a powerful open-source language model with strong reasoning capabilities. It achieves competitive performance across a wide range of natural language understanding and generation benchmarks. For the evaluation purpose, we will use it, as LLaMA-3 is designed to be efficient, making it suitable for both academic and industrial use. Fine-tuning LLaMA-3 enables domain adaptation, instruction-following enhancement, and low-resource language modeling, making it a versatile foundation model for downstream applications. Phi-3 is Microsoft's recent model specifically designed for code and mathematical reasoning. It is compact yet highly capable, making it well-suited for resource-efficient deployment. Fine-tuning Phi-3 enables specialization for domain-specific programming languages, formula derivation, or structured task automation and offers a robust foundation for applications in education, software engineering, and scientific computing.

\subsection{Dataset}

For our experiments, we utilize the GSM8K (Grade School Math 8K) dataset, which consists of 8,500 high-quality, linguistically diverse grade school math word problems created by human problem writers. The dataset is segmented into 7,500 training problems and 1,000 test problems. These problems require multi-step reasoning, typically taking between 2 and 8 steps to solve, with solutions primarily involving a sequence of elementary calculations using basic arithmetic operations (addition, subtraction, multiplication, and division).

\subsection{Baselines}

For our evaluation, we compare our proposed method with several baseline approaches:
\begin{enumerate}
    \item  \textbf{Full fine-tuning}, which utilizes the entire dataset to establish an upper performance bound. This approach represents the maximum achievable performance when all available data is used for model training.

    \item  \textbf{Random Subset Selection}, which involves randomly selecting B examples from the entire dataset, where B represents the specified budget constraint. This simple approach serves as a fundamental baseline to assess the effectiveness of more sophisticated selection strategies, where each instruction from the set has an equal chance to be selected.

    \item  \textbf{Determinantal Point Process (DPP)}, which employs a greedy approach to select diverse subsets of data. DPP explicitly models the diversity among selected examples by using determinants of kernel matrices.

    \item \textbf{Our Method}, which utilizes perplexity as a utility measure and cosine similarity as a diversity score.
\end{enumerate}

Table 1 presents the comparison of accuracy results across different subset sizes for all methods. The experiments were conducted using the Phi-3 language model with full fine-tuning for 1 epoch across all methods.
\begin{table}[h]
\centering
\begin{tabular}{|c|c|c|c|}
\hline
\textbf{Subset Size} & \textbf{Random Selection} & \textbf{DPP (Greedy)} & \textbf{Our Method} \\
\hline
900 & 0.41 & 0.49 & 0.42 \\
\hline
1000 & 0.41 & 0.46 & 0.47 \\
\hline
1100 & 0.40 & 0.44 & 0.40 \\
\hline
1300 & 0.42 & 0.45 & 0.47 \\
\hline
1500 & 0.43 & 0.47 & 0.46 \\
\hline
1700 & 0.48 & 0.43 & 0.45 \\
\hline
1900 & 0.44 & 0.45 & 0.50 \\
\hline
\end{tabular} 
\vspace{0.5em}
\caption{Comparison of accuracy across different subset selection methods}
\end{table}

The results demonstrate that both DPP and our proposed method generally outperform random selection, particularly at smaller subset sizes, highlighting the importance of strategic data selection for efficient fine-tuning. The DPP approach shows particularly strong performance at the 900 example budget (0.49 accuracy), outperforming both random selection (0.41) and our method (0.42) at this budget level. However, our method shows competitive performance at the 1000 example level with 0.47 accuracy, suggesting that different selection strategies may be optimal at different budget constraints.



\subsection{Evaluation Metrics}

For evaluation of the performance the metrics that helps are given as follows:
\textbf{Accuracy}, the percentage of correctly solved problems on test sets. \textbf{Reasoning Quality}, to assess the quality of the generated reasoning steps, we are using ROUGE and BERTScore.

\begin{ack}
    This research was supported by the Department of Computer Science and Engineering at the Indian Institute of Technology, Bombay. We thank Prof. Ganesh Ramakrishnan, professor of Optimization in Machine Learning, for his valuable guidance and insights throughout this project. We are also grateful to Suraj Racha, Teaching Assistant for the subject, for his continuous support and direction. We extend our appreciation to the BharatGen team at IIT Bombay for providing the computational resources necessary for this project.
\end{ack}

\bibliographystyle{plainnat}
\bibliography{references}
\nocite{*}

\appendix

\section{Variants of Submodular Function}
\label{appendix:submodfunctions}
\begin{enumerate}
    \item \textbf{Facility Location:} \quad $f(X) = \sum_{i \in \mathcal{V}} \max_{j \in X} s_{ij}$
    \item \textbf{Mutual Information:} \quad $f(\mathcal{A}; \mathcal{D}_T) = \sum_{i \in \mathcal{D}} \max_{j \in \mathcal{A}} s_{ij} 
+ \eta \sum_{j \in \mathcal{A}} \max_{i \in \mathcal{D}_T} s_{ij}$
    \item \textbf{Conditional Gain:} \quad $f(\mathcal{A} \mid \mathcal{D}_E) =
\sum_{i \in \mathcal{D}} \max \left( \max_{j \in \mathcal{A}} s_{ij}
- \nu \max_{k \in \mathcal{D}_E} s_{ik},\, 0 \right)$.
    \item \textbf{Graph Cut:} \quad $f(X) = \sum_{i \in \mathcal{V}, j \in X} s_{ij} - \lambda \sum_{i, j \in X} s_{ij}$
    \item \textbf{Log Determinant:} \quad $f(X) = \log \det(S_X)$
\end{enumerate}

\end{document}